%% file: paper.tex
\documentclass[letterpaper, 10pt, journal, twoside]{IEEEtran}

\usepackage{graphics} 
\usepackage{epsfig} 
\usepackage{times} 
\usepackage{amsmath} 
\usepackage{amssymb}  
\usepackage{hyperref}

\input{custom_packages}
\input{custom_definitions}

\title{Blind as a bat: audible echolocation on small robots} 

\author{Frederike Dümbgen \quad Adrien Hoffet \quad Mihailo Kolundžija \quad Adam Scholefield \quad Martin Vetterli
  \thanks{This work was supported in part by ``SNF-SESAM -- Sensing and Sampling: Theory and Algorithms.'' n$^\circ$ 200021\_181978/1.}
  \thanks{All authors are with the School of Computer and Communication Sciences, EPFL, 1015 Lausanne, Switzerland. Corresponding author: Frederike Dümbgen, \textit{frederike.duembgen@gmail.com}.}

}

\begin{document} 

\maketitle 

\input{_sections/abstract}


\input{_sections/intro}
\input{_sections/related_work}

\input{_sections/methods}

\input{_sections/experiments}
\input{_sections/results}
\input{_sections/discussion}

\vspace{-0.3em}
\section*{Acknowledgments}
We would like to thank Daniel Burnier for providing the \textit{e-puck} robot and technical support, the students Isaac, Kaourintin and Emilia for their valuable help, and Karen, Eric and Mia for their feedback on the manuscript.

\bibliographystyle{IEEEtran}
\bibliography{IEEEabrv,RAL2022}

\end{document}

%% file: custom_packages.tex
\usepackage[utf8]{inputenc} 
\usepackage{graphicx}
\usepackage{footnote}
\makesavenoteenv{tabular}
\makesavenoteenv{table}

\usepackage[ruled]{algorithm2e}

\usepackage{amsthm}
\usepackage{amsfonts}
\usepackage{mathtools}

\usepackage{enumitem}[nosep]
\usepackage{siunitx}
\usepackage{units}

\usepackage{tabularx}
\usepackage{booktabs}

\usepackage{hyperref}  
\usepackage[capitalize,noabbrev]{cleveref}  
\crefformat{equation}{(#2#1#3)}
\hypersetup{%
    colorlinks=true,
    linkcolor=gray,
    urlcolor=gray,
    citecolor=gray,
    linktocpage=true
}
\usepackage{epstopdf}

\usepackage{multirow}
\usepackage{wrapfig}

\usepackage{algorithmic}
\usepackage{array}
\newcounter{rowcntr}[table]
\usepackage[font={small,it}]{caption}
\usepackage{placeins} 

\usepackage[normalem]{ulem}

\usepackage[percent]{overpic}
\usepackage{hyphenat}

\usepackage{todonotes}
\usepackage{bm}
\usepackage{textcomp}

%% file: custom_definitions.tex
\graphicspath{{drawings/}{export/}}

\newcommand{\vc}[1]{\ensuremath{\bm{#1}}}

\DeclarePairedDelimiter\abs{\lvert}{\rvert}
\DeclarePairedDelimiter\norm{\lVert}{\rVert}
\makeatletter
\let\oldabs\abs
\let\oldnorm\norm
\def\abs{\@ifstar{\oldabs}{\oldabs*}}
\def\norm{\@ifstar{\oldnorm}{\oldnorm*}}
\makeatother

\newcommand{\R}[1]{\ensuremath{\boldsymbol{\mathbb{R}}^{#1}}}
\newcommand{\inR}[1]{\ensuremath{\in \R{#1}}}

\newcommand{\SE}[1]{\ensuremath{SE({#1})}}

\theoremstyle{definition}

\let\oldunderbrace\underbrace
\renewcommand{\underbrace}[2]{\let\scriptstyle\textstyle \oldunderbrace{#1}_{#2}}
\definecolor{tab10blue}{rgb}{0.12156863, 0.46666667, 0.70588235}

\def\ca{\emph{ca}.}

\def\eg{\emph{e.g}.} 
\def\ie{\emph{i.e}.}

\DeclareFontFamily{U}{mathx}{\hyphenchar\font45}
\DeclareFontShape{U}{mathx}{m}{n}{ <-> mathx10 }{}
\DeclareSymbolFont{mathx}{U}{mathx}{m}{n}
\DeclareFontSubstitution{U}{mathx}{m}{n}
\DeclareMathAccent{\widebar}{\mathalpha}{mathx}{"73}
\newcolumntype{C}[1]{>{\centering\let\newline\\\arraybackslash\hspace{0pt}}m{#1}}
\newcolumntype{D}[1]{>{\centering\let\newline\\\arraybackslash\hspace{0pt}\columncolor{gray!20}}m{#1}}
\newcolumntype{R}[1]{>{\raggedleft\let\newline\\\arraybackslash\hspace{0pt}}m{#1}}
\newcolumntype{L}[1]{>{\raggedright\let\newline\\\arraybackslash\hspace{0pt}}m{#1}}
\newcolumntype{M}[1]{>{\raggedright\let\newline\\\arraybackslash\hspace{0pt}}m{#1}}
\newcolumntype{N}{>{\refstepcounter{rowcntr}\therowcntr}c}

\newcommand{\mic}{\ensuremath{^{(m)}}}
\newcommand\pose{\ensuremath{\bm{\xi}_n}}
\newcommand\planel{\ensuremath{\bm{\pi}_l}}

\usepackage{tabstackengine}
\stackMath
\setstackgap{L}{18pt}
\setstacktabbedgap{0pt}
\def\lrgap{\kern6pt}

\newcommand{\changed}[1]{#1}
\newcommand{\english}[1]{#1}

%% file: _sections/abstract.tex
\begin{abstract}
For safe and efficient operation, mobile robots need to perceive their environment, and in particular, perform tasks such as obstacle detection, localization, and mapping.
Although robots are often equipped with microphones and speakers, the audio modality is rarely used for these tasks. Compared to the localization of sound sources, for which many practical solutions exist, algorithms for active echolocation are less developed and often rely on hardware requirements that are out of reach for small robots.  

We propose an end-to-end pipeline for sound-based localization and mapping that is targeted at, but not limited to, robots equipped with only simple buzzers and low-end microphones. 
The method is model-based, runs in real time, and requires no prior calibration or training.
We successfully test the algorithm on the e-puck robot with its integrated audio hardware, and on the Crazyflie drone, for which we design a reproducible audio extension deck. We achieve centimeter-level wall localization on both platforms when the robots are static during the measurement process. Even in the more challenging setting of a flying drone, we can successfully localize walls, which we demonstrate in a proof-of-concept multi-wall localization and mapping demo.
\end{abstract}

%% file: _sections/intro.tex
\section{Introduction}



\IEEEPARstart{A}{} \english{bat's} ability to ``see'' in the dark using sound has fascinated scientists for centuries~\cite{Fenton1992}. Bats emit ultrasonic chirps, and based on the timing and form of the echoes, localize objects of interest such as food~\cite{Simon2011}, obstacles~\cite{Griffin1941}, or water resources~\cite{Greif2010}.

Similarly, most mobile robots need to localize targets and obstacles in their surroundings in order to perform meaningful operations such as search and rescue, exploration\english{,} and delivery. Most robots that are \english{currently} deployed in the real world rely on rich sensors for localization and mapping --- sensors \english{that} provide a plethora of information --- such as cameras, lidar~\cite{Cadena2016} or radar~\cite{Lu2020}. 

However, just like bats have relatively poor eyesight, not all robots can be equipped with rich sensing equipment. Consider \english{the} \textit{Crazyflie}, a developer-friendly nano drone~\cite{Crazyflie}.
This drone is very limited in terms of admissible payload, \english{which rules out} heavy sensors such as lidar or radar. A single camera can be attached through the \textit{AI-deck}; however, this does not provide full coverage for obstacle avoidance. More coverage can be obtained with the \textit{multi-ranger deck}, designed specifically for obstacle detection using four infrared sensors.

We propose to instead use audible sound for obstacle detection and mapping. 
\english{Microphones come in lighter and smaller form factors than rich sensors and demand less memory for processing.} Compared to proximity sensors such as infrared and ultrasound, they are less directional, allowing fewer sensors to achieve full spatial coverage. 
\english{Moreover, many robots that are designed to communicate with users come equipped with microphones, and} thus using audio for navigation tasks may require little additional payload.

\english{In} this paper, we develop a novel interference-based echolocation algorithm, which includes a simultaneous localization and mapping (SLAM) framework, integrating a particle filter and a factor graph~\cite{Dellaert2021}, which accounts for the nonlinear, multi-modal nature of echolocation measurements. For experimental evaluation, we develop and provide an audio extension deck for the \textit{Crazyflie} drone. \english{It consists of a small piezoelectric buzzer and four microphones embedded in a custom PCB, including its own microcontroller for acquisition and preprocessing.}
The methods developed in this paper are however applicable to any robot equipped with at least one microphone and a speaker. To underline this point, we also provide experimental validation on the ground-based \textit{e-puck} robot~\cite{epuck}. Both platforms are shown in Figure~\ref{fig:setups}. To summarize, our main contributions are: 

\begin{figure}[tb]
  \centering
  \includegraphics[width=\linewidth]{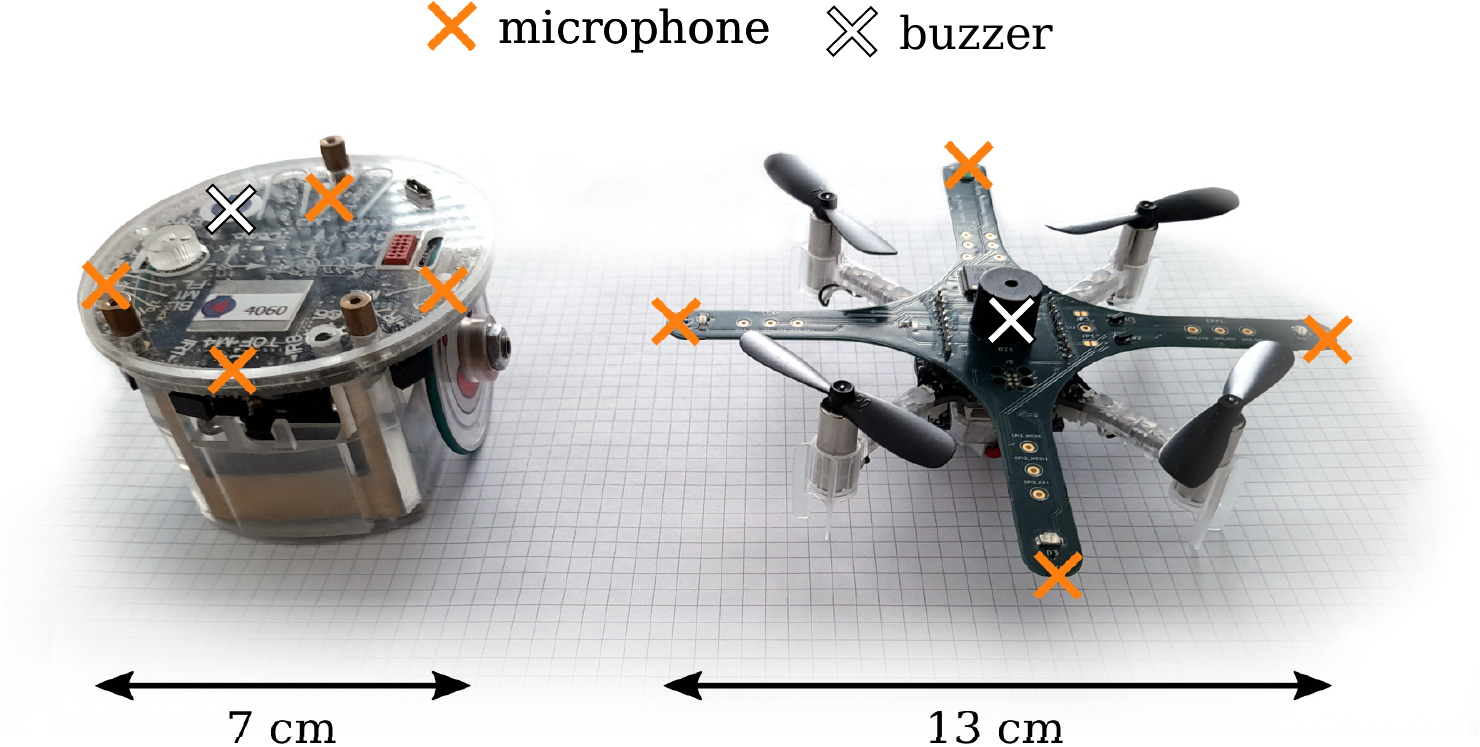}
  \caption{The robotic platforms used for echolocation in this paper. Depicted on the left is the \textit{e-puck} robot~\cite{epuck}, a wheeled robot developed for education. On the right, we show the \textit{Crazyflie} drone~\cite{Crazyflie} with our custom audio extension deck, equipped with four microphones and a piezoelectric buzzer. The proposed echolocation pipeline can be used on any platform with at least one low-key microphone and sound source, respectively.}\label{fig:setups}
\end{figure}

\begin{itemize}
  \item a model-based\english{,} probabilistic echolocation algorithm based on audio interference measurements, 
  \item a combined particle filtering and SLAM pipeline suited for the non-Gaussian nature of echolocation measurements, and 
  \item a \textit{Crazyflie} audio extension deck with associated software for research on audio-based navigation on drones.
\end{itemize}

The paper is structured as follows. After reviewing related work in Section~\ref{sec:related-work}, we describe our method for going from low-level audio signals to wall and robot location estimates in Section~\ref{sec:methods}. In Section~\ref{sec:hardware}, we provide implementation details for the two used hardware platforms. In Section~\ref{sec:results}, we evaluate the method on these platforms -- with finely controlled motion in Section~\ref{sec:results-stepper}, on the flying drone with one wall in Section~\ref{sec:results-flying}, and in a multi-wall demo in Section~\ref{sec:results-demo}. We conclude with a discussion of the results in Section~\ref{sec:discussion}.

%% file: _sections/related_work.tex
\section{Related Work}\label{sec:related-work}

We first give an overview of existing solutions for echolocation from raw audio signals, also called the front-end in the SLAM literature~\cite{Cadena2016}, before going over relevant work in the back-end state estimation procedure.
\paragraph{\textbf{Sound-based front end}} Existing methods for \changed{sound-based sensing} can be loosely categorized \english{into} passive solutions, which \english{estimate} the direction of arrival (DOA) of external sound sources, and active solutions, which probe the environment with an onboard sound source. \changed{While DOA for robotics is a rather active research field~\cite{DOAreview}, sound-based active solutions, the focus of this paper, are considerably less explored.} 

\changed{Amongst} the most popular active solutions are methods estimating the time of arrival (TOA) of the echoes of reflecting objects. In rooms, this information is contained in the so-called room impulse response (RIR), consisting of shifted, attenuated peaks for the early echoes, and a more noise-like reverberation tail. It can be estimated in frequency domain through the deconvolution of a known wide-band source signal and its response. When the locations of microphones and speakers are known, the first peaks of the RIR can be used to approximately recover a room's geometry~\cite{Dokmanic2013}. Adapting this method to a moving setup, the authors of~\cite{Krekovic2016} map out the room, one wall at a time, using a small loudspeaker and high-quality microphones mounted on a wheeled robot. The authors of~\cite{Saqib2020} use a similar platform but recover the peaks of the RIR through nonlinear least-squares optimization on the wide-band frequency response. In follow-up work~\cite{Saqib2022}, this method is extended with a DOA estimator. 
Although promising, these methods use loudspeakers and microphones with high signal-to-noise ratio (SNR) and flat frequency responses, which is not available on the platforms studied in this paper. 

\changed{The} TOA of reflectors can also be determined by emitting and cross-correlating short, frequency-modulated pulses, a concept known as pulse compression. Using this technique, room geometry is estimated, using a static omnidirectional speaker and a tablet's microphones, in~\cite{Shih2019}. Moving to a fully mobile setup, the authors of~\cite{Zhou2020} perform room mapping on a smartphone. Pulse compression is also commonly used for underwater acoustic ranging~\cite{Rypkema2017,Webster2012}. Note that these methods require fine control of the emitted audio pulses.

\english{This last point is overcome by methods exploiting sound interference}. \english{The idea} is to detect close-by reflectors by measuring the change in magnitude caused by interference from its echoes. In the first work bringing this idea to robotics, it \english{was shown} that a reflector can be detected based on the change in magnitude of the propeller's ego-noise~\cite{Calkins2020}. The reported results are obtained using static measurement microphones, and an offline calibration phase of the propeller's ego noise. \changed{The results highlight one advantage of interference-based methods: since they do not require emission of specifically designed short pulses, they can be used on noise inherent to the system.}
 
\changed{In this paper, we take an interference-based approach. Our system is lighter compared to RIR-based methods,} both in terms of hardware requirements and computing power: it runs on computationally limited platforms with simplistic speakers and low-key microphones. \changed{Compared to pulse-compression techniques, the sound signal is simpler and does not need to be exactly known --- this can be an advantage when the available buzzer does not allow for fine-grained control, or when sound sources inherent to the system are used. As opposed to~\cite{Calkins2020}, our system does not require prior calibration, uses on-board low-key microphones and tightly integrates the motion of the device, as discussed next.} 

\paragraph{\textbf{\changed{Back-end state estimation}}} \changed{For a practical solution, the measurements taken at subsequent times and poses need to be fused into a coherent map. This can be posed as a factor graph inference problem, where all unknown states (poses and walls) are linked by measurement factors~\cite{Dellaert2021}. Inference on factor graphs can become prohibitively expensive as the number of unknown states increases over time, unless certain Gaussian and sparsity assumptions are made, which is exploited by modern solvers~\cite{kaess_isam_2008,Kaess2012,Barfoot2020}. While the Gaussian assumption may hold for many commonly used sensors, it does not for the sound-based front end considered in this paper: sound is known to lead to multi-modal distributions~\cite{Cheung2019,Fourie2020} and thus requires different inference approaches~\cite{Fourie2016,FourieThesis}. Our state estimation pipeline is most similar in nature to the work by~\cite{Rypkema2017}, where a particle filter is used as a pre-filtering step to render the bearing and range pseudo-distributions approximately Gaussian. In our case, the particle filter helps not only reduce ambiguities, but also allows us to convert the raw path difference measurements to distance and angle estimates, which can be fed directly to the SLAM algorithm.}

%% file: _sections/methods.tex
\section{Methods}\label{sec:methods}

\begin{figure*}[t]
  \centering
  \includegraphics[width=.9\linewidth]{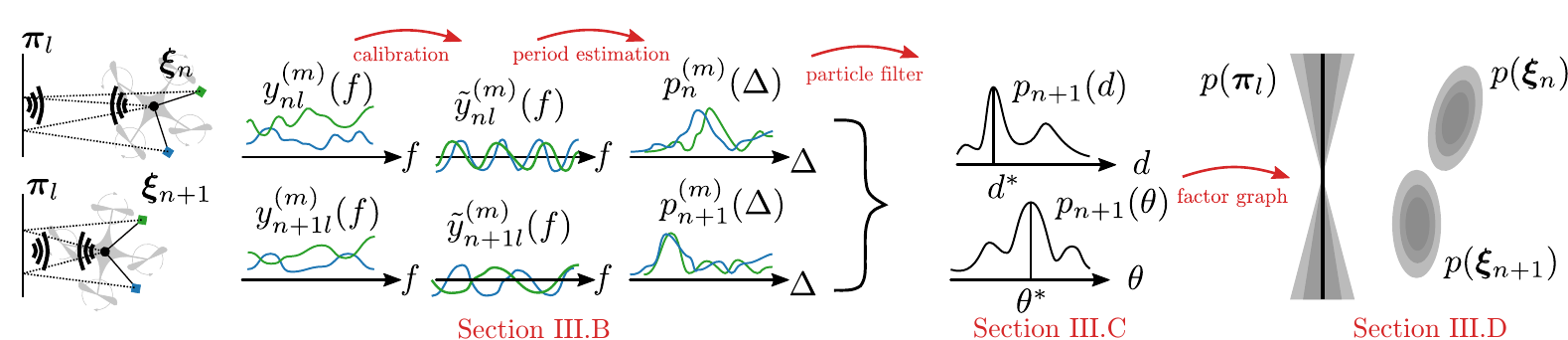}
  \caption{ \changed{Overview of estimation pipeline: we depict two microphones and their measurements in green and blue, respectively, at two subsequent poses. The raw measurements $y_{nl}\mic(f)$, shown on the left, are used to estimate the location of nearby walls $p(\vc{\pi}_l)$ and the robot's poses over time $p(\vc{\xi}_n)$, shown on the right. }}
  \label{fig:pipeline}
\end{figure*}

In this section, we describe the estimation pipeline, going from raw audio measurements to pose and plane estimates. We first derive the model of the microphone measurements as a function of the robot pose and plane in Section~\ref{sec:signal-model}. We derive our measurement model, which can be used to estimate the path difference, in Section~\ref{sec:interference-model}, and \changed{infer} distance and angle distributions using discrete filtering in Section~\ref{sec:filtering}.  We conclude with a standard SLAM framework to jointly estimate poses and planes in Section~\ref{sec:slam}. \changed{An overview of the estimation pipeline is given in Figure~\ref{fig:pipeline}.}

\subsection{Derivation of signal model}\label{sec:signal-model}

 
We first derive the model of the individual microphone responses. A sketch of a sample setup is shown Figure~\ref{fig:notation}. We place the robot's body frame (also called the local frame) at the sound source location $\vc{s}\inR{2}$, and denote the $N_m$ microphone locations in this frame by $\vc{x}_m\inR{2}$. The pose, consisting of the robot's translation $\vc{x}_n\inR{2}$ and orientation $\phi_n\in[0, 2\pi)$ at time $t_n$, is denoted by $\vc{\xi}_n$, expressed in the fixed inertial frame (also called global frame). \english{Considering the studied experimental platforms (a ground-based robot and a drone hovering at constant height) we operate in two dimensions}, but this is not a hard requirement for the proposed solution. We denote the $N_l$ planes by $\vc{\pi}_l$, and we parametrize them with their global distance $d_l$ and angle $\theta_l$ from the origin. 
The recorded signal at each microphone is the sum of the direct sound and the reflected sound.  Accounting for free-space attenuation and energy dissipation according to~\cite{Borish1984}, it takes the form:
\begin{align}
    &z_{nl}\mic(t) := z\mic(t, \pose, \planel) \label{eq:signal-time}  \\
    &= g\mic\left( \frac{1}{4\pi \ell\mic}s\bigg(t-\frac{\ell\mic}{c}\bigg) + \frac{1-\rho}{4 \pi r\mic_{nl}}s\bigg(t-\frac{r\mic_{nl}}{c}\bigg)\right),\nonumber
\end{align}
where $c$ is the speed of sound, $\rho$ the wall absorption coefficient, and $g\mic$ the unknown, frequency-dependent, microphone gain. The \english{lengths of the direct and reflected paths are respectively} given by $\ell\mic := \| \bm{x}\mic - \bm{s} \|$, and
\begin{equation}
  r_{nl}\mic = \sqrt{{\left(\ell\mic\right)}^2 + 4{\left(d_{nl}\right)}^2 - 4d_{nl}\ell\mic \cos(\theta\mic - \theta_{nl})},
  \label{eq:reflected-path}
\end{equation}
\noindent where $\theta\mic=\angle{(\vc{x}\mic - \vc{s})}$ denotes the bearing of the $m$-th microphone, and $\theta_{nl}$, $d_{nl}$ denote the distance and angle of plane $l$ as seen from pose \pose: 
\begin{equation}
  d_{nl} = d_l - \vc{x}_n^\top{\vc{n}\big(\theta_l\big)} \quad \text{and} \quad \theta_{nl} = \theta_l - \theta_n,
  \label{eq:distance-angle-local}
\end{equation}

\begin{center}
  \centering
  \includegraphics[width=\linewidth,draft=false]{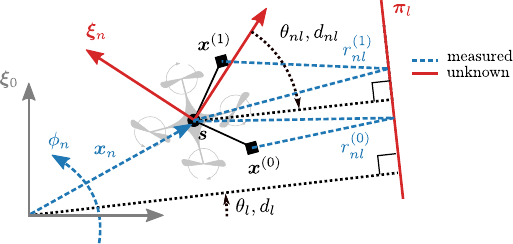}
  \captionof{figure}{
  Sketch of the experimental setup of echolocation with one wall, two microphones and \changed{one} source. We can measure the difference between the reflected path lengths $r_{nl}\mic$ for mics $m=0, 1$, and the direct path lengths $\ell\mic$, respectively, which are related to the distance $d_{nl}$ and angle $\theta_{nl}$ of the wall $\vc{\pi}_l$ at pose $\vc{\xi}_n$. 
  }\label{fig:notation}
\end{center}

\noindent
where $\vc{n}(\theta)$ is a normal vector with direction $\theta$.  Solving~\eqref{eq:reflected-path} for the distance $d_{nl}$ given a fixed angle, we have:
\begin{align}
  d_{nl} &:= h\left(r_{nl}\mic, \theta_l\right) = \frac{1}{2}\bigg( \ell\mic \cos{(\theta\mic-\theta_{nl})} \bigg. \label{eq:g-func} \\
  & \bigg. + \sqrt{{\left(\ell\mic\right)}^2\left(\cos^2{(\theta\mic-\theta_{nl})} - 1\right) + \left(r_{nl}\mic\right)^2}\bigg),\nonumber
\end{align}
which, for the limit case \english{of a} co-located phone and speaker simplifies to the widely used approximation $d_{nl} \approx r_{nl}\mic / 2$.

\subsection{\english{Interference: from audio to path differences}}\label{sec:interference-model}

Next, we derive our measurement function. In \changed{the} frequency domain, the squared magnitude of~\eqref{eq:signal-time}  becomes:
\begin{equation}\label{eq:model-full2}
  \begin{aligned}
    y_{nl}\mic(f) 
    &= \frac{{g\mic}{(f)}^2|\hat{s}(f)|^2}{{(4\pi)}^2} \Bigg( \frac{1}{{\ell\mic}^2} + \frac{{(1-\rho)}^2}{{\left(r_{nl}\mic\right)}^2} \Bigg. \\
    & \Bigg.+ 2\frac{1-\rho}{\ell\mic r_{nl}\mic}\underbrace{\cos{\left(2\pi f(r_{nl}\mic - \ell\mic)/c\right)} }{:=\tilde{y}_{nl}\mic(f)}\Bigg),
  \end{aligned}
\end{equation} 
where we introduce $\Delta_{nl}\mic:=r_{nl}\mic - \ell\mic$ for the path difference. In theory, $y_{nl}\mic(f)$ could be used as the measurement model: evaluating it at $N_f$ different frequencies $f_k$, the robot state \pose~and wall location~\planel~\english{can} be inferred using nonlinear optimization. However, the function is highly nonlinear in the unknowns and, in particular, non-bijective, making it prone to yield ambiguous results. To make things worse, the unknown parameters $g\mic(f)$, $|\hat{s}(f)|$ and $\rho$ in~\eqref{eq:model-full2} render the optimization problem more costly, should we try to either estimate them (thus adding more dimensions to the state vector), or marginalize them out. 

\changed{Thankfully}, since the ``frequency''\footnote{This ``frequency'' has the unit of seconds, and is not to be confused with the ``real'' frequencies $f_k$ (expressed in Hz) at which we measure.}  of the cosine in~\eqref{eq:model-full2} depends only on the unknown state vectors and \english{sound} frequency, we can infer probability distributions of \english{the path difference $\Delta\mic$ by studying its frequency response}. We introduce $\tilde{y}_{nl}\mic(f)$ for the cosine, as shown in~\eqref{eq:model-full2}. It can be \changed{estimated} from $y_{nl}\mic(f)$ through online calibration, which we will further develop in Section~\ref{sec:calibration}.\footnote{An interesting result states that the FFT-based solution we propose is in fact equivalent to least-squares optimization with amplitude and phase of the cosine marginalized out~\cite{Vanderplas2018}.}

\begin{figure}[h]
  \centering
  \vspace{0.5em}
  \includegraphics[width=.9\linewidth,draft=false]{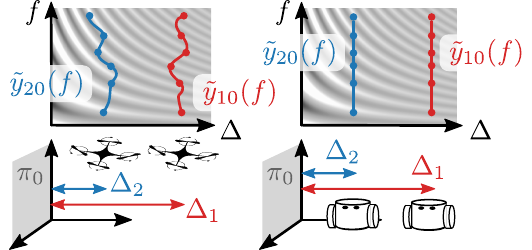}
  \caption{\changed{Sketch of interference patterns over path differences and frequencies, with the samples recorded by a drone (left) and a ground-based robot (right). For different distances, we measure different periodicities in the interference. Note that the ground-based robot can measure clean vertical slices, as opposed to the drone.}}\label{fig:matrices}
\end{figure}

Using a \english{classical result from} Bayesian theory, we know that the probability of a signal to be a sinusoid of frequency $f$ is directly related to the signal's periodogram~\cite{Jaynes1987}. \changed{Applying this result here, the probability $p_n\mic(\Delta)$ of microphone $m$ being at a path difference $\Delta$ at time $n$ is given by:}
\changed{
\begin{align}
  \label{eq:prob-delta}
p_n\mic(\Delta) &= \eta {\left[1 - \frac{2 P\mic_{\tilde{y}}(\Delta/c)}{\sum_{k=1}^{N_f}{\big(\tilde{y}_{nl}\mic(f)\big)^2}} \right]}^{\frac{2-N_f}{2}}, \\
    \text{with }P_{\tilde{y}}\mic(f) &:= \frac{1}{N_f} \left| \sum_{k=1}^{N_f} \tilde{y}_{nl}\mic(f) e^{j2\pi f f_k}\right|^2,
  \label{eq:periodogram}
\end{align}
where $\eta$ is a normalization factor.}  By choosing a roughly uniform spacing of frequencies, we can use the FFT of $\tilde{y}$ to calculate $P_{\tilde{y}}\mic(f)$. 
\changed{For visualization purposes, we sketch an example of $\tilde{y}_{nl}\mic(f)$, for different frequencies and distances, in Figure~\ref{fig:matrices}. Intuitively speaking, we sample an approximately vertical slice \changed{of this function}, and the periodicity of the slice is related to the path difference of the closest wall.}

Next, we fuse multiple path difference distributions to obtain distance and angle distributions that can be fed to \changed{our} SLAM framework. 

\begin{figure*}[t]
  \centering
  \includegraphics[width=\linewidth,draft=false]{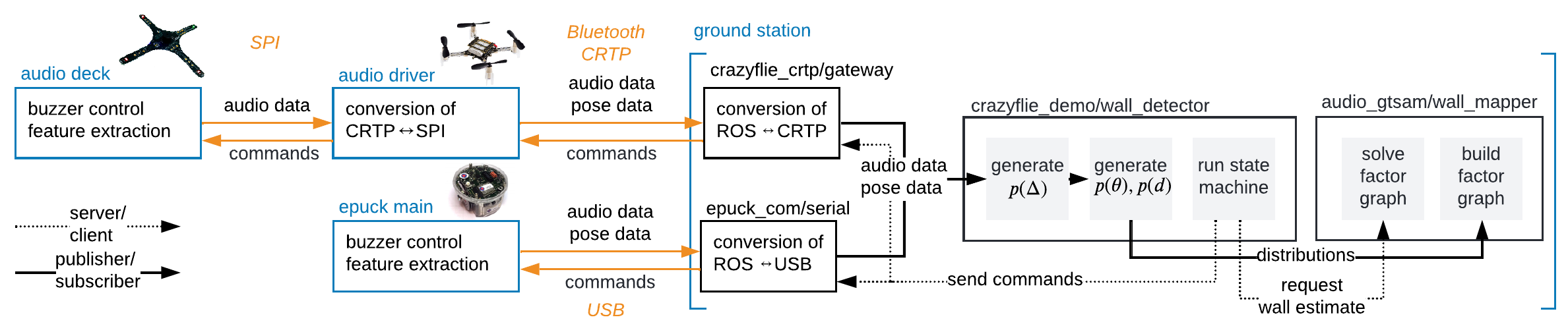}%
  \vspace{-1em}
  \caption{Processing pipeline from firmware to the ground
    station's ROS processing pipeline. The audio data and pose data is sent from either the \textit{Crazyflie} drone (top row) or the \textit{e-puck} robot (bottom row) to the ground station, where they are processed in parallel by a series of ROS nodes. \changed{Note that service/client connections (dotted lines) are only triggered upon request, while listener/publisher connections (solid black lines) are running continuously}. 
}\label{fig:software}
\end{figure*}

\subsection{\changed{Particle filter}: from path differences to planes}\label{sec:filtering}


Because of the non-linear measurement model and the poor signal quality \english{that} common light-weight microphones and buzzers provide, the path difference distributions are subject to high noise levels and ambiguities. We propose a particle filtering approach for converting them to more unimodal \english{distance and angle distributions}. The proposed filter \english{implicitly assumes} that, for a small time window, the pose estimates are subject to only little drift, and can thus be used to resolve the ambiguities in the distributions. Mathematically speaking, we want to convert our potentially ambiguous path distributions $p\mic_n(\Delta)$, obtained from~\eqref{eq:prob-delta}, to unimodal distance- and angle distributions $p(d_l)$ and $p(\theta_l)$. 






\paragraph{Initialization}

We initialize $N_p$ particles, containing the (local) angle and distance candidates $(d_{0,k}, \theta_{0, k})$\english{, with $k=0\ldots N_p-1$}.  The distances and angles are chosen uniformly from an area of size $[0, d_{\max}] \times [0, 2\pi)$. We denote by $d_{\max}$ the maximal distance that we aim to estimate, given by the maximally resolvable path difference.\footnote{By the Nyquist criterion we have $d_{\max} = h(\Delta_{\max} + \ell) = h(c / 2\delta_f + \ell) \approx c/4\delta_f$, with $\delta_f$ the (average) difference between the sampled frequencies.} Because of low SNR at high distances, we clip this maximal distance at \SI{80}{cm}.  We initialize all particles with equal weights $w_k = 1 / N_p$. 

\paragraph{Prediction} At each timestep, we use the current movement estimates to transform the particles to the new local coordinates: 
\begin{equation}
  \begin{aligned}
  d_{n,k} &= d_{n-1,k} - {\vc{n}(\theta_{n-1,k} + \phi_{n})}^\top(\vc{x}_n - \vc{x}_{n-1}) \\
  \theta_{n,k} &= \theta_{n-1,k} - (\phi_n - \phi_{n-1}), \\
  \end{aligned}
  \label{eq:prediction}
\end{equation}
\changed{\noindent where $\phi_n$ and $\vc{x}_n$ are relative pose estimates.}
We add a fixed proportion (10\%) of uniformly distributed particles, in order to prevent the filter from converging to local minima.

\paragraph{Update} \english{Next}, we update the particle weights using the newest measurements $p\mic_n(\Delta)$. Using each particle's $(d_{n,k}, \theta_{n,k})$ values, \english{we} calculate the corresponding path difference $\Delta\mic_k$ with~\eqref{eq:reflected-path} and \english{evaluate the distributions using linear interpolation.} \changed{Assuming the microphone measurements are independent, we set the value $w_k$ to the product of the $N_m$ probabilities thus obtained.}

\paragraph{Resampling} After each update step, we resample the particles according to their weights, using the stratified resampling algorithm~\cite{Sarndal2003}. 

\subsection{\changed{Planar} simultaneous localization and mapping}\label{sec:slam}

In the final step, we aim to feed the filtered plane measurements, along with potential pose measurements, to a SLAM framework. This ensures long-term consistency, and enables crucial tasks such as data association and path planning. 

In the factor graph representation of our problem, the nodes correspond to poses and planes. The pose estimates $\tilde{\vc{\xi}}_n\in\SE{d}$ are incorporated in unary factors of the form:
\begin{equation}
  c_n = \norm{\vc{\xi}_n \ominus \tilde{\vc{\xi}}_n}_{\Sigma_n},
  \label{eq:pose-factor}
\end{equation}
where $\ominus$ is the difference operator defined in the Lie algebra of \SE{d}, and $\Sigma_n$ is the diagonal pose covariance matrix, fixed heuristically according to the pose estimation accuracy of the state estimator ($\sigma_x=\sigma_y=\SI{1}{cm}$, $\sigma_\phi=\SI{5}{\deg}$). 

For the plane measurements, we first extract distance and angle estimates $\tilde{d}_{nl},\tilde{\theta}_{nl}$ \changed{and standard deviations $\sigma^{\theta}_{nl}, \sigma^d_{nl}$ from the particle filter}, using the weighted particle mean and standard deviation. Approximating the distance and angle distributions as Gaussians, we use the binary plane factors of the form~\cite{Trevor2012}: 
\begin{equation}
  c_{nl} = \norm{\begin{bmatrix} \vc{R}_n^\top \vc{n}(\theta_l)  \\
    \vc{x}_n^\top \vc{n}(\theta_l)\end{bmatrix}
  - \begin{bmatrix}\vc{n}(\tilde{\theta}_{nl}) \\ \tilde{d}_{nl}  \end{bmatrix}}_{\Sigma_{nl}},
  \label{eq:plane-factor}
\end{equation}
\noindent where $\vc{R}_n$ is the rotation matrix corresponding to $\phi_n$. $\Sigma_{nl}$ is the noise covariance matrix, calculated from $\sigma^{\theta}_{nl}, \sigma^d_{nl}$, with the upper-left block given by:
\begin{equation}
  ({\sigma^{\theta}_{nl}})^2 \frac{\partial\vc{n}}{\partial \theta}\Big|_{\tilde{\theta}_{nl}} \frac{\partial\vc{n}}{\partial \theta}\Big|_{\tilde{\theta}_{nl}}^\top.
  \label{eq:sigma}
\end{equation}

To determine whether a new measurement corresponds to a new plane or to a revisited plane, we introduce the following data association loss between planes $\vc{\pi}_i$ and $\vc{\pi}_j$: 
\begin{equation}
  e(\vc{\pi}_i, \vc{\pi}_j) = \norm{\vc{n}(\theta_i)d_i - \vc{n}(\theta_j)d_j}.
  \label{eq:data-association}
\end{equation}
This loss measures how far the normal points of the two planes are from each other, thus incorporating both angle as well as distance \english{discrepancies}. We associate a new plane measurement to a previously seen plane if~\eqref{eq:data-association} falls below a given threshold. We found that a threshold of \SI{30}{cm} performed well throughout our experiments. 
We solve the  factor graph using the \textit{iSAM2} solver implemented in \textit{gtsam}~\cite{Kaess2012}.

%% file: _sections/experiments.tex
\begin{figure*}[ht] 
  \vspace{1em}
  \centering 
  \centering 
  \begin{minipage}{4cm}
    \centering
    \includegraphics[draft=false,trim={0cm 0.5 15cm 10cm},clip,width=3.7cm]{_figures/stepper-setup-close.pdf} \\
    \vspace{0.5em} 
    \includegraphics[draft=false,trim={3cm 3cm 5cm 1cm},clip,height=3.7cm,angle=90]{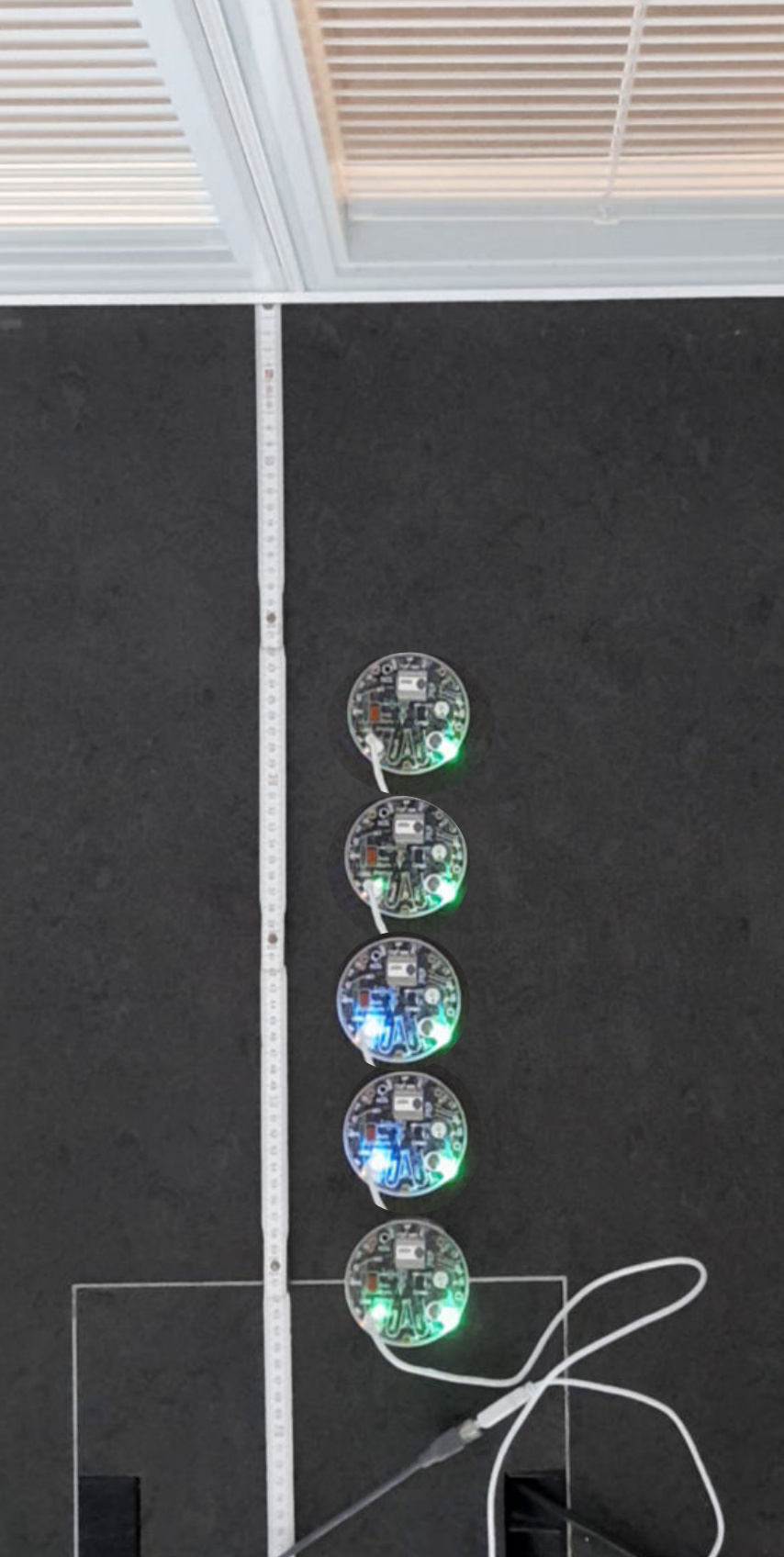}
  \end{minipage}
  \begin{minipage}{14cm}
    \centering
  \includegraphics[draft=false,width=.49\linewidth]{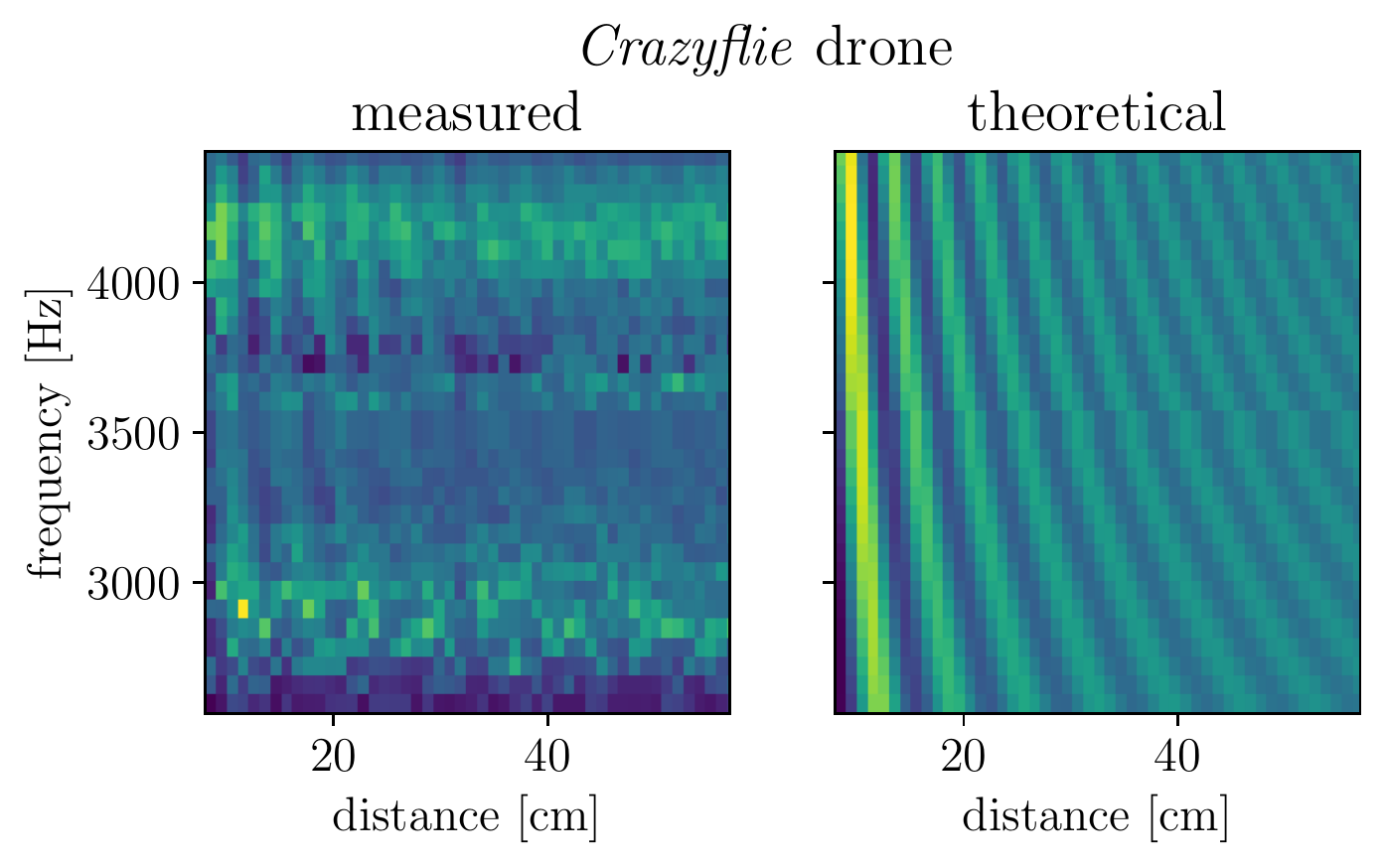} 
  \includegraphics[draft=false,width=.49\linewidth]{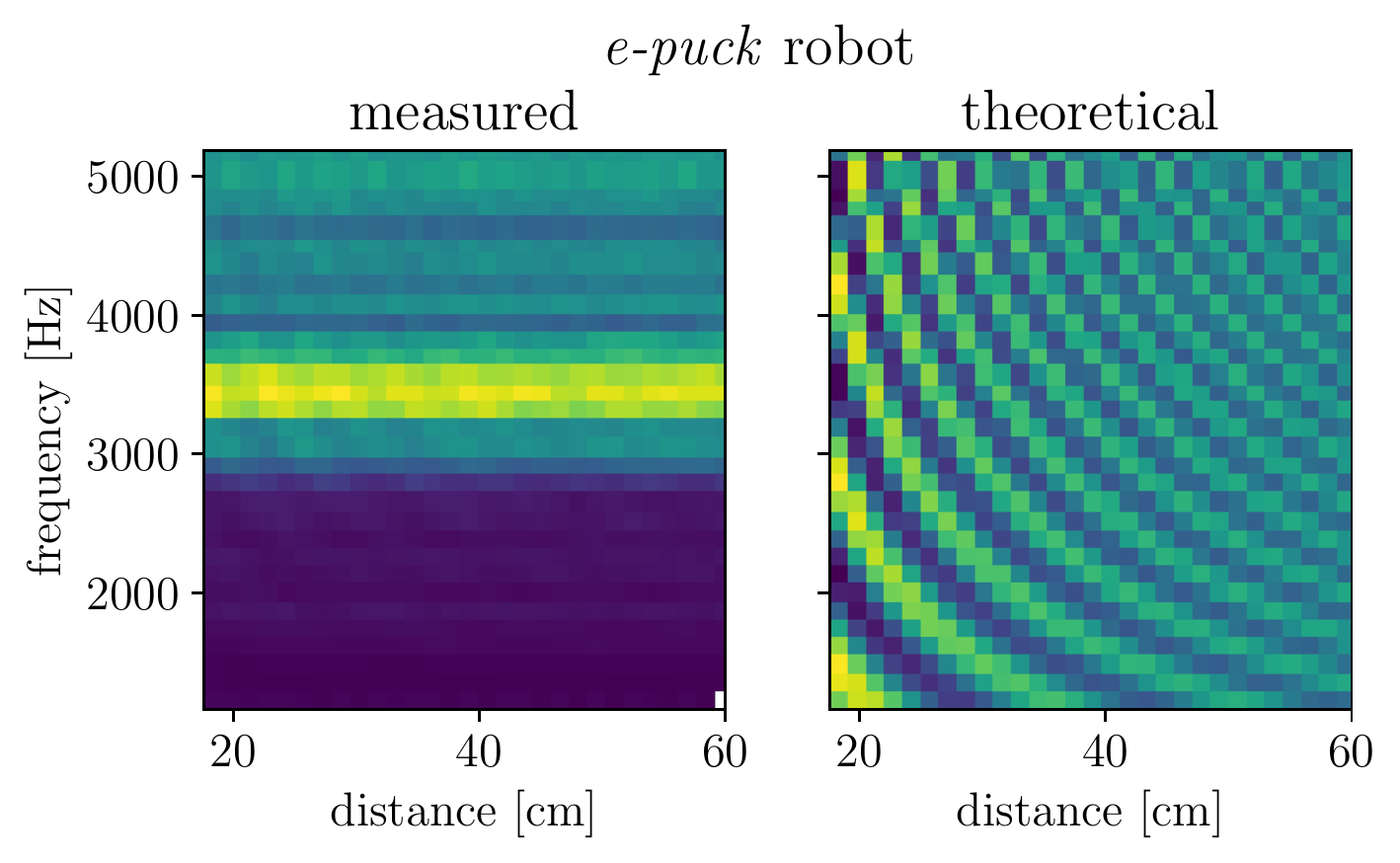}
\end{minipage}
    \caption{\changed{Interference patterns measured at one microphone,} using the drone's stepper-motor
    setup and the \textit{e-puck} setup. Photos of the experimental setups are shown to the left. \changed{Note that the measured values have the same patterns as predicted by theory, using equation~\eqref{eq:model-full2}. }}\label{fig:results-matrices} 
\end{figure*}

\section{Experimental platforms}\label{sec:hardware}

To demonstrate the described algorithm's performance, the main experimental platform studied is the \textit{Crazyflie} drone, a micro-drone with
relatively poor computing resources (microcontroller STM32F405) and strict payload
constraints. To fit these requirements, we design an audio-deck optimized to add
little weight, and equipped with its own microcontroller (STM32F446) to offload
the low-level audio pre-processing. The deck conforms with the standard format of
\textit{Crazyflie} extension decks, thus adhering to its plug-and-play philosophy.
The deck contains four MEMS microphones attached to its extremities, a
positioning chosen for its good trade-off between minimal propeller noise and
off-centre weight. A piezoelectric buzzer, which has a reduced PCB footprint and energy consumption 
compared to a speaker, is placed in the centre of the deck. 
Note that one microphone is enough for the described algorithms, but
having more microphones both increases the SNR and enables 
other applications such as DOA estimation. Relative pose estimates are obtained using the drone's optical flow deck.
A photo of the \textit{Crazyflie} equipped with the audio deck is shown in Figure~\ref{fig:setups}, and all hardware design files are made publicly available.\footnote{The code base including hardware design files and ROS pipeline is available at \url{https://github.com/LCAV/audioROS}.} 

We also test our algorithms on the wheeled \textit{e-puck} robot (version 2)~\cite{epuck}.  Just like the custom audio
deck, the robot possesses four MEMS microphones, but it has a more versatile and louder electromagnetic speaker.
Being a ground-based robot, the \textit{e-puck} allows for the evaluation of the
methods in more controlled settings, eliminating both position jitter and
propeller noise. Relative pose estimates are obtained through wheel odometry.

\subsection{Low-level audio processing}

The full processing pipeline is visualized in Figure~\ref{fig:software}. In the firmware (left part of the plot), the signals recorded by the four microphones are first stored in buffers of size $N=2048$ at a sampling rate of
$F_s=\SI{48}{kHz}$. With these values, the resulting frequency resolution, using
the real-valued FFT, is $\Delta_f = F_s / N \approx \SI{23}{Hz}$. 

\changed{The} piecewise constant frequency sweep of the buzzer is generated through PWM signals. \changed{For the drone, to increase SNR, the frequency range is chosen as to be minimally affected by propeller noise under average drone hovering thrusts, while ensuring that the played frequencies match the available frequency analysis bins.} 
Each note is played until the microphones have recorded one full buffer, which takes
$N / F_s \approx \SI{40}{ms}$. For a sweep of $N_f = 20$ discrete frequencies,
this results in circa~\SI{1}{s} per sweep. After each sweep, the complex response
is calculated through a windowed FFT, using the flattop window,
which yields faithful magnitude measurements even under small frequency errors~\cite{flattop}.
The response at the current buzzer frequency is extracted and added to an  
outgoing buffer. 

After completion of each sweep, the next sweep is started only
once the data was successfully sent to the ground station. For the \textit{Crazyflie} drone, the buffer is passed first via
SPI to the main processor and then through the custom Bluetooth protocol (CRTP) to the ground station. The data throughput of both SPI and CRTP being
significantly higher than the playback rate of the sweeps $N_f \times 2
\times 32 = \SI{1280}{bps}$, we can ensure smooth transmission. For the \textit{e-puck} robot, we omit the intermediate SPI and CRTP communication and send the audio data directly over USB to the ground station. 


\subsection{Ground station processing} 

On the ground station (right half of Figure~\ref{fig:software}), the received audio data is
passed through a modular processing pipeline implementing the
steps described in Sections~\ref{sec:interference-model} to~\ref{sec:slam} through custom nodes and
interfaces, written in \textit{python} in the \english{Robot} Operating Systems (ROS) framework, \changed{version \textit{Galactic}}.  
The full software suite, including a simulation framework and implementations of DOA methods, are made publicly available. 

\subsection{Online joint calibration}\label{sec:calibration}

In our model~\eqref{eq:model-full2}, there are two unwanted and unknown, frequency-dependent gains: the highly non-linear frequency response of the piezo buzzer $|\hat{s}(f)|$, and the gains of the different microphones $g\mic(f)$. If we remove these two factors, we can use the measurement model $\tilde{y}_{nl}\mic(f)$, and path difference estimation is reduced to a simple frequency analysis, as described in~\ref{sec:interference-model}. Since the two unknown gains are entangled, we choose to estimate them jointly. The intuition of the calibration is the following: while the robot is moving, it measures a portion of the distance-frequency function, as visualized in Figure~\ref{fig:matrices}. Even for small movements, \ie~small variations of wall distance $d$, we are likely to sample both destructive and constructive interference regimes. We can thus expect that taking a moving average over a sufficiently long time window yields a good estimate of the joint speaker-microphone gain. To reduce the computational cost of the calibration, we implement the average using a simple infinite impulse response (IIR) low-pass filter:
\begin{equation}
  \tilde{g}_n\mic(f) 
  = (1 - \lambda) \tilde{g}_{n-1}\mic(f) + \lambda y_{nl}\mic(f),
  \label{eq:calibration}
\end{equation}
\noindent where we have introduced $\tilde{g}_n\mic(f)$, the estimate of the joint gain function $g\mic(f)^2|\hat{s}(f)|^2$ at time index $n$. The parameter $\lambda$ controls the effective window width, and we use a value of $\lambda=0.3$ throughout all experiments. 

%% file: _sections/results.tex
\section{Results}\label{sec:results}

We evaluate our systems in three increasingly difficult stages. First, \english{we obtain fine-grained measurements of the wall's interference patterns by moving the drone and \textit{e-puck} robot in small steps towards the wall. For this purpose, the drone is mounted on a linear stepper motor.} In a second stage, we evaluate the drone's performance in localizing a \english{single wall during flight, before concluding with a multi-wall demo}.

\input{_sections/results-stepper.tex}

\subsection{Results on flying drone}\label{sec:results-flying} 

\begin{figure}[h!]
  \centering
\includegraphics[width=\linewidth]{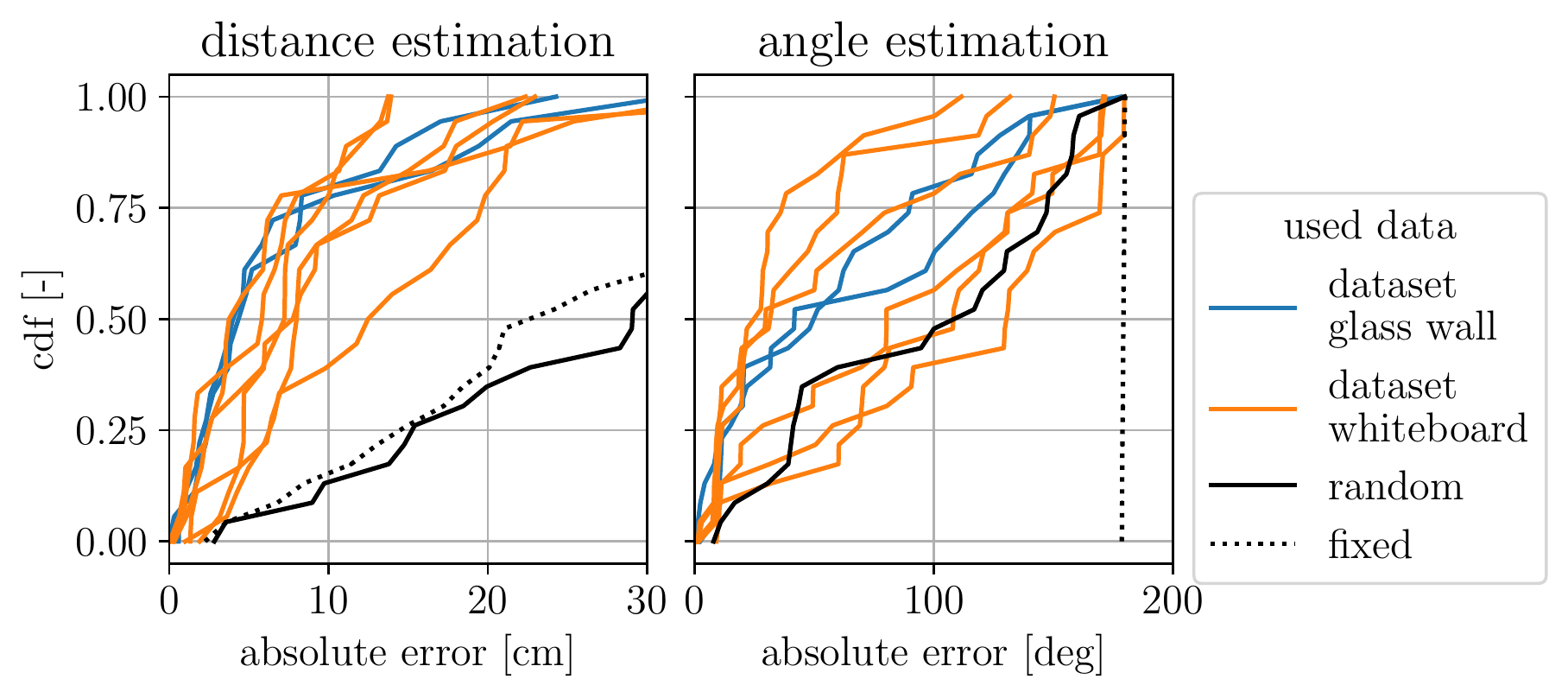} \\
\includegraphics[angle=90,height=2.2cm,draft=false]{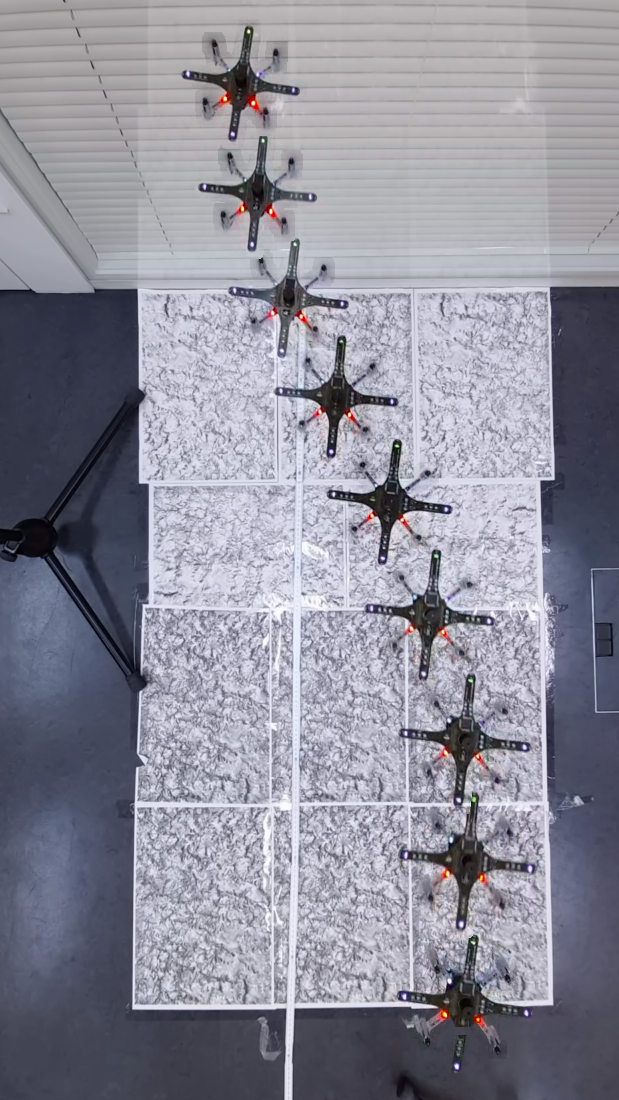} 
  \includegraphics[height=2.2cm,draft=false,trim={2cm 8cm 14cm 6cm},clip]{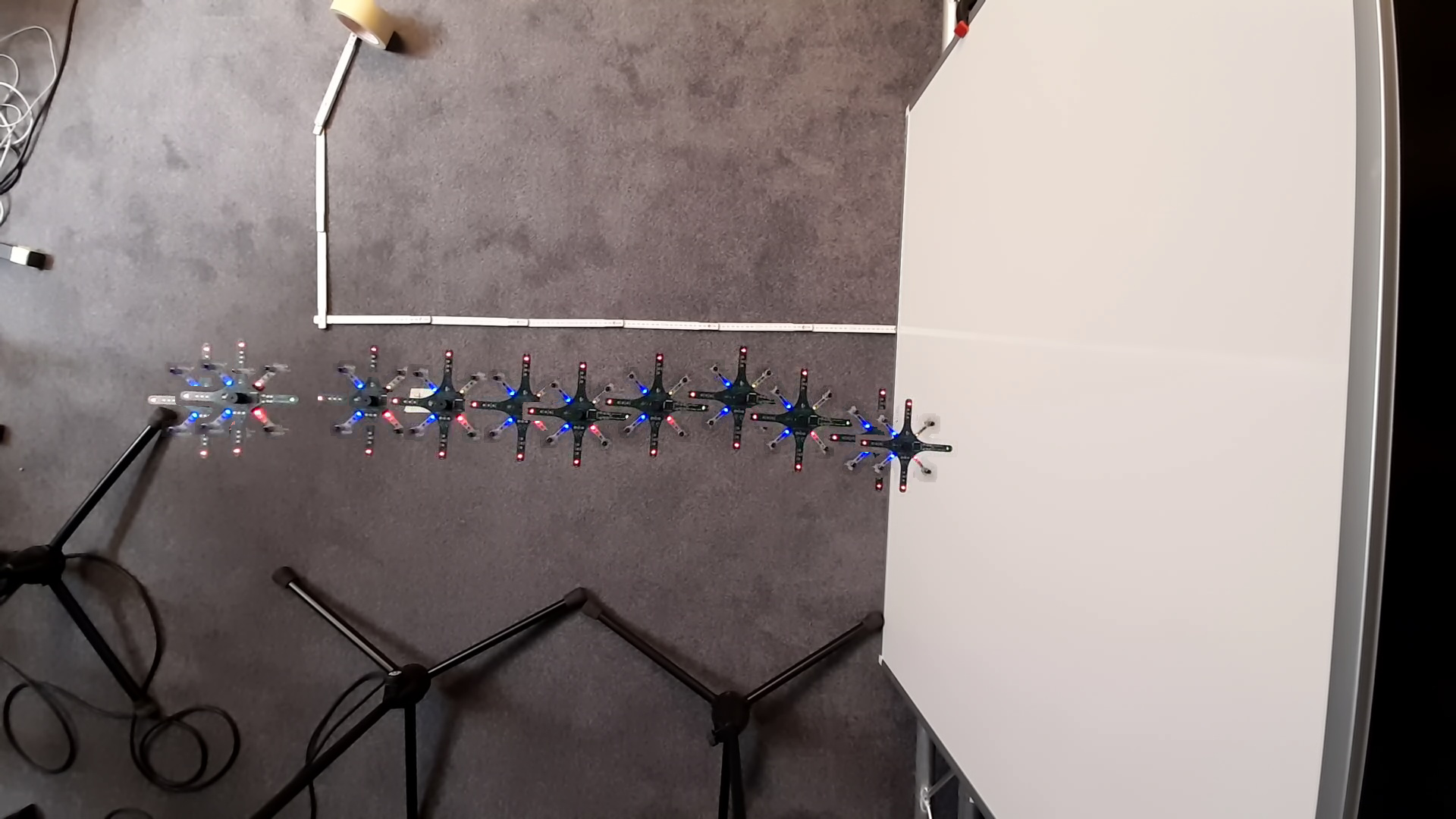}
  \caption{Distance and angle estimation by the flying drone approaching the wall. \changed{Top: cumulative distribution functions of the absolute errors, where each line corresponds to a different experiment. The same baselines as for Figure~\ref{fig:results-stepper-cdfs} are used}. Bottom: \english{merged} consecutive frames of two example experiments (bird view).}\label{fig:results-flying3}
\end{figure}

\changed{The results thus far suggest that the proposed algorithm works well on robots with precise motion. For the remainder of this paper, we move to the more challenging setup of a flying drone.} We \english{fly the drone
at constant speed towards a flat surface (glass or whiteboard), of which we estimate the location}. This
setup adds to the above studies the nuisance factors of 
1) lateral movement during measurements and 2) varying propeller noise. To reduce both, we pick a shorter frequency sweep of $N_f=20$ and omit the softer middle frequencies. The results are more variable, as shown in Figure~\ref{fig:results-flying3}. The drone however still localizes the wall with a median distance error of less than \changed{\SI{8}{cm}} for almost all test runs. The angle estimates, on the other hand, \changed{lose in accuracy, with performances ranging from\SI{30}{deg} up to \SI{120}{deg} median error}. One possible explanation is that the movement between two consecutive sweeps is relatively small, making angle estimation very sensitive to noise.  

\subsection{Multi-wall demo}\label{sec:results-demo} 
\begin{figure}[h!]
  \centering
  \vspace{0.5em}
  \includegraphics[width=\linewidth, draft=false]{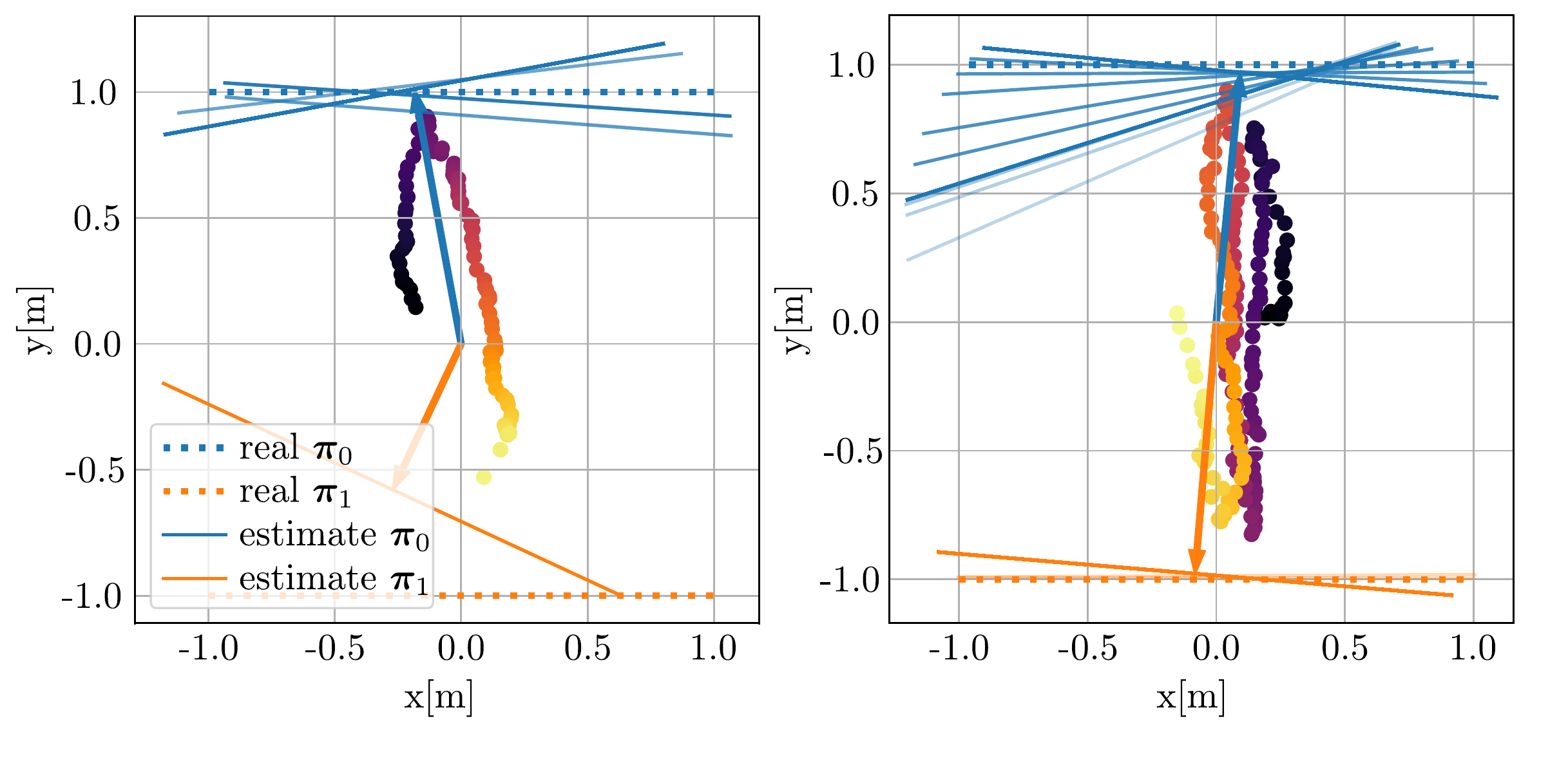} \\
  \includegraphics[width=\linewidth, trim={0cm 9cm 3cm 5cm}, clip, draft=false]{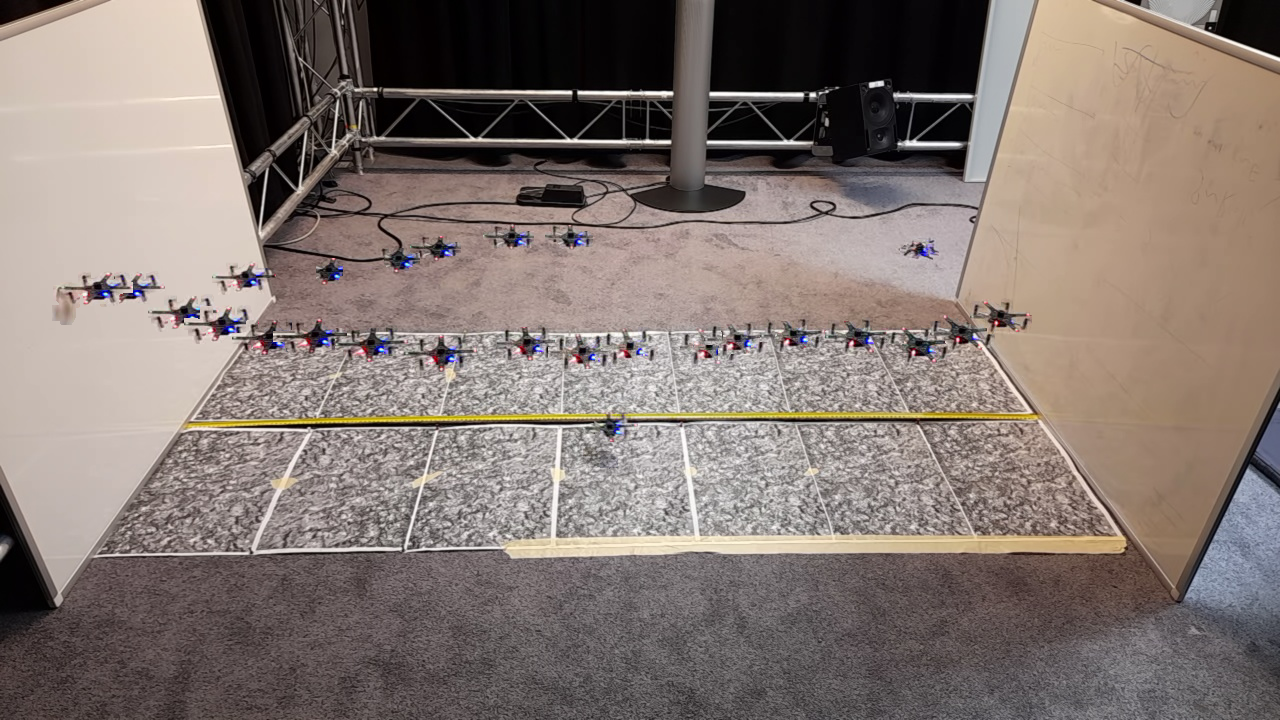}  \\
  \caption{\changed{Live wall avoidance experiments. Top: Estimated poses and planes after factor-graph inference, for two experiments. The positions over time are plotted in yellow to dark-blue. Wall estimates are plotted with decreasing transparency over time. Bottom: merged consecutive frames of first experiment (side view).}
  }\label{fig:demo}
\end{figure}

Finally, we implement a multi-wall detection demo. The drone flies back and forth between two whiteboards using only sound to detect, avoid, and map them, as shown in Figure~\ref{fig:demo}. We control the drone manually, but the processing runs in real time on the recorded \textit{bag} files.

We assume there is a wall ahead of the drone when the distance estimate is below a threshold of \SI{20}{cm}. Once a wall is detected, we perform data association by~\eqref{eq:data-association}. Because of the low accuracy of angle estimates observed in the flying experiments, we use the simplification that \english{a detected wall is always in front of the drone with respect to its movement}.  
The average processing time of the non-optimized \textit{python} implementation of the factor-graph inference, is less than \SI{1}{ms}, even when running on all nodes (up to 200 for the second demo). Together with the inference time from the particle filter, the state estimation can thus run at \ca~\SI{10}{Hz}.

We qualitatively evaluate the performance of the full pipeline in Figure~\ref{fig:demo}, where we show the drone's pose estimates, wall estimates and ground truth wall locations. The drone successfully maps and avoids the walls, and the labelling works reliably. In particular, in the second experiment (right plot in Figure~\ref{fig:demo}), new measurements are correctly associated with $\vc{\pi}_0$ when revisiting the first wall.

%% file: _sections/results-stepper.tex
\subsection{Stepper motor and ground-based robot}\label{sec:results-stepper} 

Moving the drone and \textit{e-puck} robot in centimeter-steps perpendicular to a glass wall, we create the measured \english{interference matrices shown in}
Figure~\ref{fig:results-matrices}. The frequency sweeps consist of $N_f=32$ uniformly spaced frequencies, where the range for the drone is chosen so as to avoid the propeller noise harmonics. \changed{We note that, although noisy, we can clearly discern the interference patterns predicted by theory}. The pattern is particularly pronounced closer to the wall. We also note a strong frequency-dependent gain, particularly for the \textit{e-puck} robot, which we remove through online calibration as described in Section~\ref{sec:calibration}.

\begin{figure}[t]
  \vspace{0.5em}
  \includegraphics[height=4.2cm, draft=false]{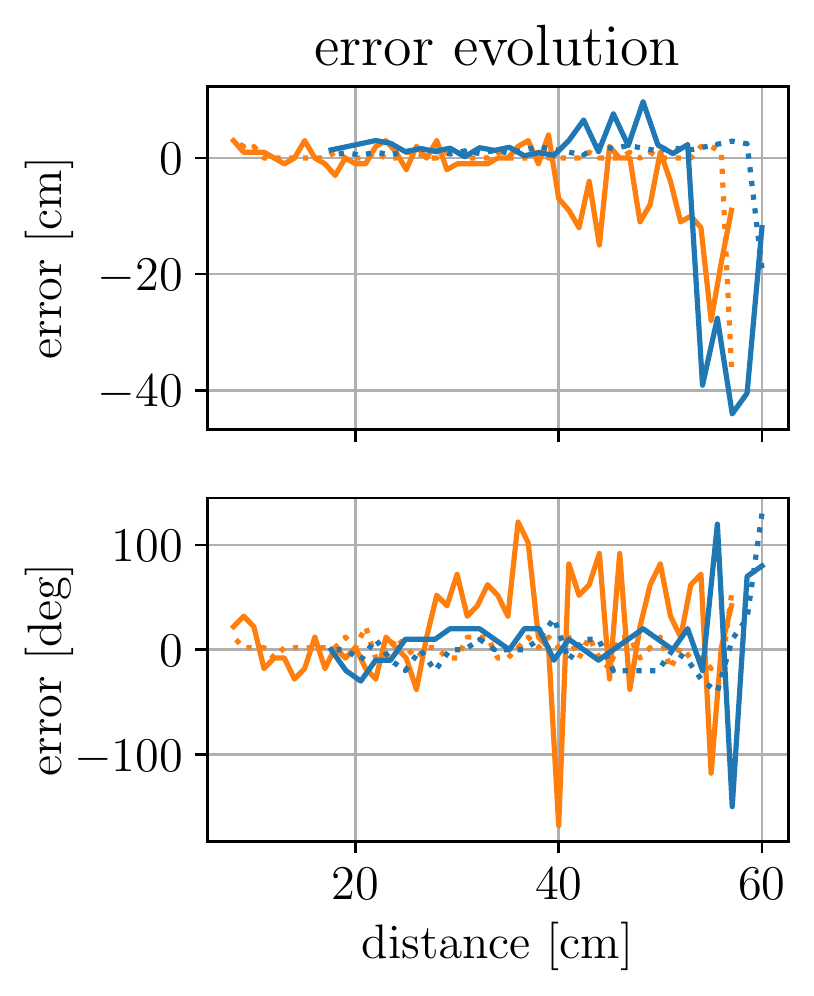}
  \includegraphics[height=4.2cm, draft=false]{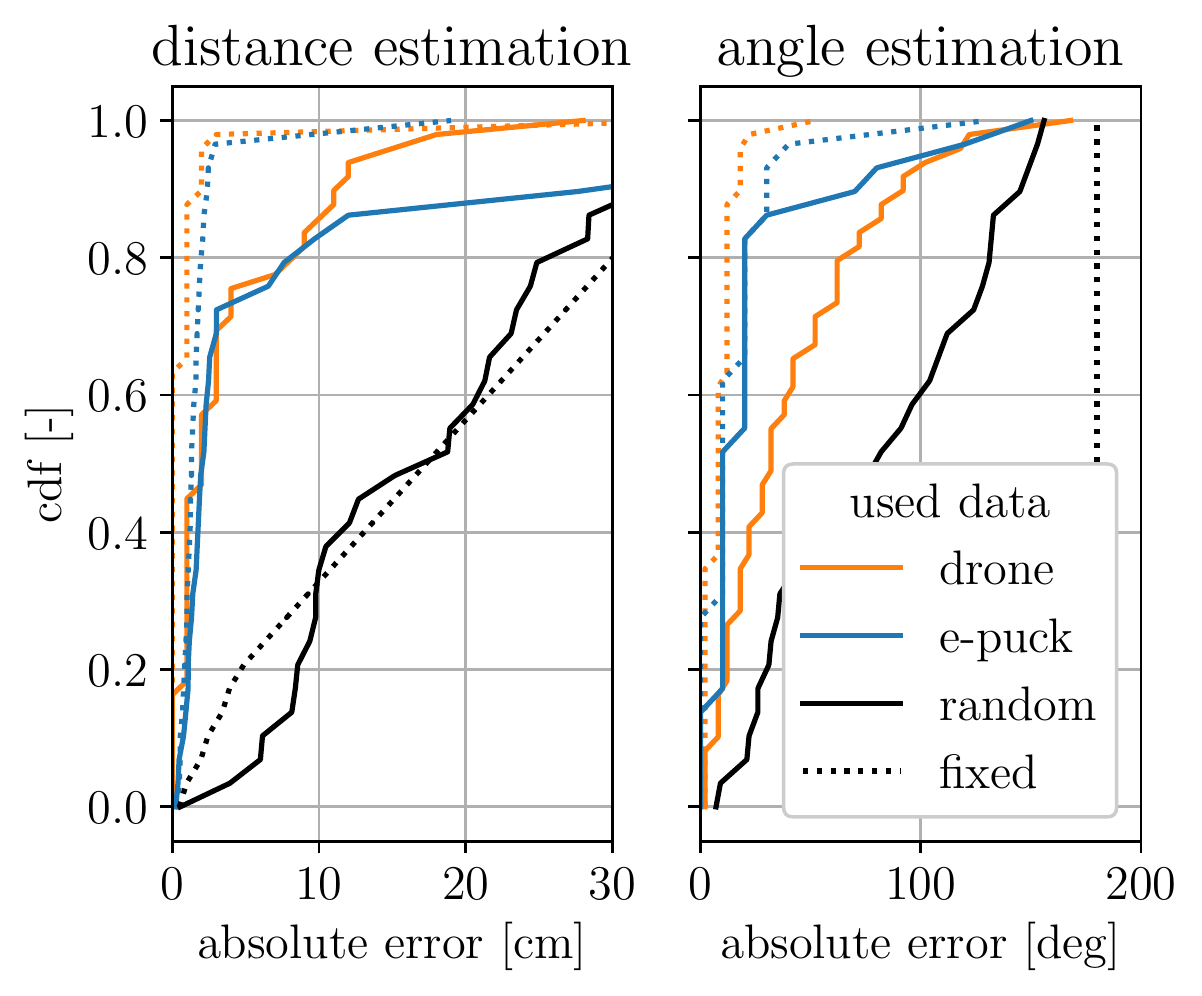}
  \caption{\changed{Performance of distance and angle estimation. Left: distance and angle error as a function of wall distance. Right: cumulative distribution functions (cdf) of absolute errors. Shown are the experimental results from the drone on the stepper motor and the \textit{e-puck} robot. For comparison, we also include the results using simulated data. The error using random guesses or one fixed guess serve as worse-case baselines. }}\label{fig:results-stepper-cdfs}
\end{figure}

\changed{Next, we apply our estimation pipeline to these measurements. Treating each frequency sweep (each column in Figure~\ref{fig:results-matrices}) separately, we evaluate the performance of both distance and angle estimation. We found that $N_p=400$ particles gave consistently good results throughout all experiments, with which one full estimation (including update, prediction and resampling) takes less than \SI{75}{ms}.
We first investigate the estimation error as we increase the distance to the wall (left plots in Figure~\ref{fig:results-stepper-cdfs}). Distance estimation is reliable up to \SI{40}{cm} from the wall, with the maximum error staying below \SI{2}{cm}. Angle estimation is less reliable, in particular for the drone. However, up to \ca~\SI{30}{cm}, the maximum error is smaller than \SI{20}{deg}. We take a closer look at the cumulative probability distributions (cdfs) in the right plots of Figure~\ref{fig:results-stepper-cdfs}. As worst-case baselines, we plot the result of \english{choosing random angles or distances}, or using a fixed value at the mid-point. As best-case baseline, we run the proposed solution on simulated data. We observe that the median error for distance estimation for both \textit{e-puck} and drone experiments is less than \SI{3}{cm}. The \textit{e-puck} robot also performs particularly well at angle estimation with a median error below \SI{10}{deg}.}

\begin{figure}[t]
  \vspace{0.5em}
  \includegraphics[width=\linewidth, draft=false]{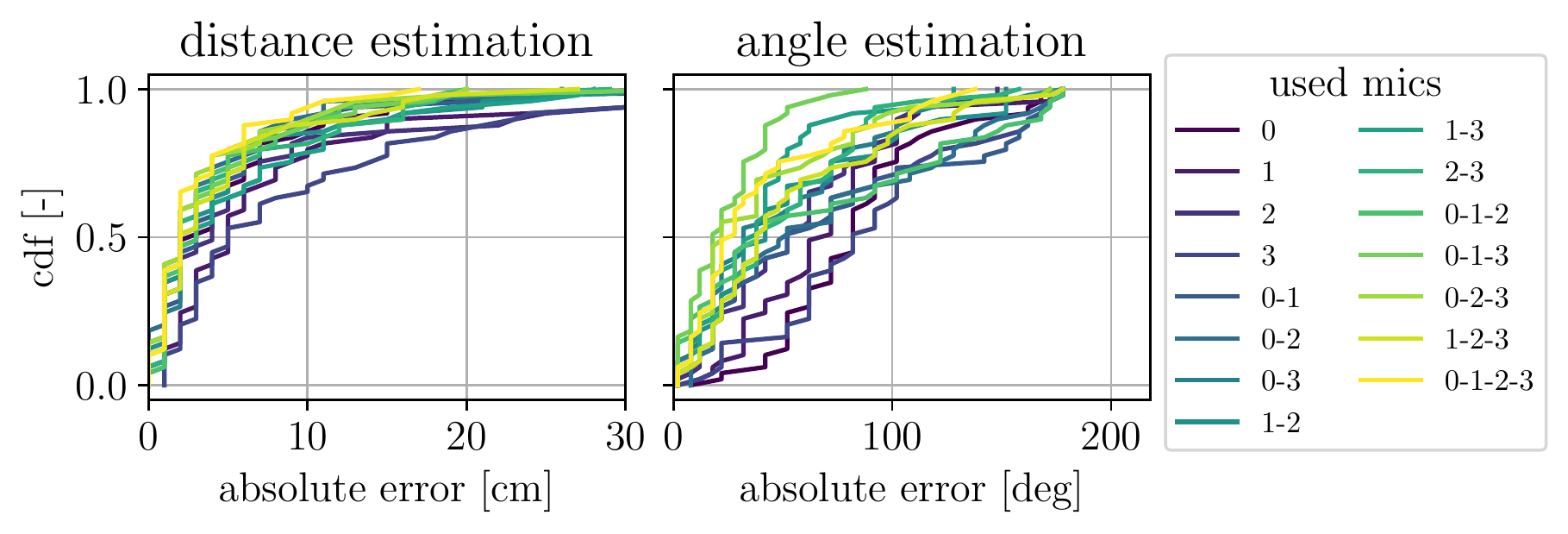}
  \caption{\changed{Ablation study for the number of microphones used for distance and angle estimation on the drone (stepper-motor setup). While performance degrades with fewer microphones, it is still satisfactory even for a single microphone.}}\label{fig:mic-ablation}
\end{figure}

\changed{Finally, we perform an ablation study on the number of microphones used for estimation. Focusing on the drone results for brevity, Figure~\ref{fig:mic-ablation} shows that the performance goes down as we exclude more microphones. However, even for a single microphone, we still achieve median errors of less than \SI{5}{cm} and \SI{90}{deg}, which is sufficient for robust wall detection.}

%% file: _sections/discussion.tex
\section{Conclusion and future work}\label{sec:discussion}

We have proposed an end-to-end system for wall localization based on audio measurements. Using our custom audio deck for the \textit{Crazyflie} drone and the off-the-shelf \textit{e-puck} robot, we show how an approximate interference model can be used to determine the location of a reflector. This algorithm is suited for light-weight and cheap hardware, and runs entirely in real time without prior calibration. 

\english{For ground-based robots or drones with a controlled movement,}
we achieve a precision of less than \SI{2}{cm} and \SI{30}{deg} median error over a distance of \SI{50}{cm}. For the more challenging setting of a flying drone, the accuracy lowers to~\ca~\SI{8}{cm}, but the method is still able to perform wall detection reliably, which we demonstrate in a multi-wall detection demo. For the latter, the estimates are fed into a factor-graph SLAM pipeline which also performs data association.

\changed{The performance drop when moving from ground-based to flying platforms suggests that the system would benefit from more sophisticated algorithms for handling propeller noise. For example, varying propeller noise and sound interference (\eg~background noise like voices or music) may affect the performance. Adaptive frequency sweep selection based on surrounding noise bears the potential to greatly increase the generalizability and robustness of the system.}
\changed{Compared} to distance estimation, \english{angle estimation was found to be more sensitive to noise.}\changed{ In future work, we plan to add other direction indicators} such as sound level or phase differences \changed{as input to the angle estimation.} Furthermore, \changed{an} implicit assumption of the proposed method is that the robot is static during each frequency sweep. While this is easily achieved for ground-based robots, a method more specifically targeted to drones \changed{could} account for \english{its} noisy motion.
\changed{Finally, to overcome the potential nuisance of audible signals, a promising direction of future research consists of replacing the buzzer signal with a system's inherent sound sources such as propeller noise.}

